\documentclass[10pt,twocolumn,letterpaper]{article}

\usepackage{cvpr}
\usepackage{times}
\usepackage{epsfig}
\usepackage{graphicx}
\usepackage{amsmath}
\usepackage{amssymb}
\usepackage{subfigure}
\usepackage{color}
\usepackage{multirow}
\usepackage{threeparttable}

\usepackage[breaklinks=true,bookmarks=false]{hyperref}

\cvprfinalcopy 

\def\cvprPaperID{****} 

\begin{document}

\title{DeepHuman: 3D Human Reconstruction from a Single Image}

\author{
	Zerong Zheng\footnotemark[1]
	\and Tao Yu\footnotemark[1]{ }{ }\footnotemark[2]
	\and Yixuan Wei\footnotemark[1]
	\and Qionghai Dai\footnotemark[1]
	\and Yebin Liu\footnotemark[1]\\
	\and \footnotemark[1]{ }{ }Tsinghua University
	\and \footnotemark[2]{ }{ }Beihang University
}

\maketitle

\begin{abstract}
We propose DeepHuman, an image-guided volume-to-volume translation CNN for 3D human reconstruction from a single RGB image. To reduce the ambiguities associated with the surface geometry reconstruction, even for the reconstruction of invisible areas, we propose and leverage a dense semantic representation generated from SMPL model as an additional input. One key feature of our network is that it fuses different scales of image features into the 3D space through volumetric feature transformation, which helps to recover accurate surface geometry. The visible surface details are further refined through a normal refinement network, which can be concatenated with the volume generation network using our proposed volumetric normal projection layer. We also contribute THuman, a 3D real-world human model dataset containing about 7000 models. The network is trained using training data generated from the dataset. Overall, due to the specific design of our network and the diversity in our dataset, our method enables 3D human model estimation given only a single image and outperforms state-of-the-art approaches.   
\end{abstract}

\section{Introduction}
Image-based reconstruction of a human body is an important research topic for VR/AR content creation \cite{Head-wornTotalCap}, image and video editing and re-enactment \cite{Skeletal-SurfaceMotionGraph,MotionGraph}, holoportation \cite{Holoportation} and virtual dressing \cite{ClothCap}. To perform full-body 3D reconstruction, currently available methods require the fusion of multiview images \cite{collet2015high,TotalCapture,SparseViewHaoLi18} or multiple temporal images \cite{VideoAvater,DetailedHumanAvater} of the target. Recovering a human model from a single RGB image remains a challenging task that has so far attracted little attention.  Using only a single image, available human parsing studies have covered popular topics starting from 2D pose detection \cite{DeepCut,OpenPoseCao,Hourglass16}, advancing to 3D pose detection \cite{SimpleYetEffective,Rogez17,Zhou17}, and finally expanding to body shape capture \cite{HMR} using a human statistic template such as SMPL \cite{SMPL:2015}. However, the statistic template can capture only the shape and pose of a minimally clothed body and lack the ability to represent a 3D human model under a normal clothing layer. Although the most recent work, BodyNet\cite{BodyNet}, has pioneered research towards this goal, it only generates nearly undressed body reconstruction results with occasionally broken body parts. We believe that 3D human reconstruction under normal clothing from a single image, which needs to be further studied, will soon be the next hot research topic.

\begin{figure}
	\centering
	\includegraphics[width=1.0\linewidth]{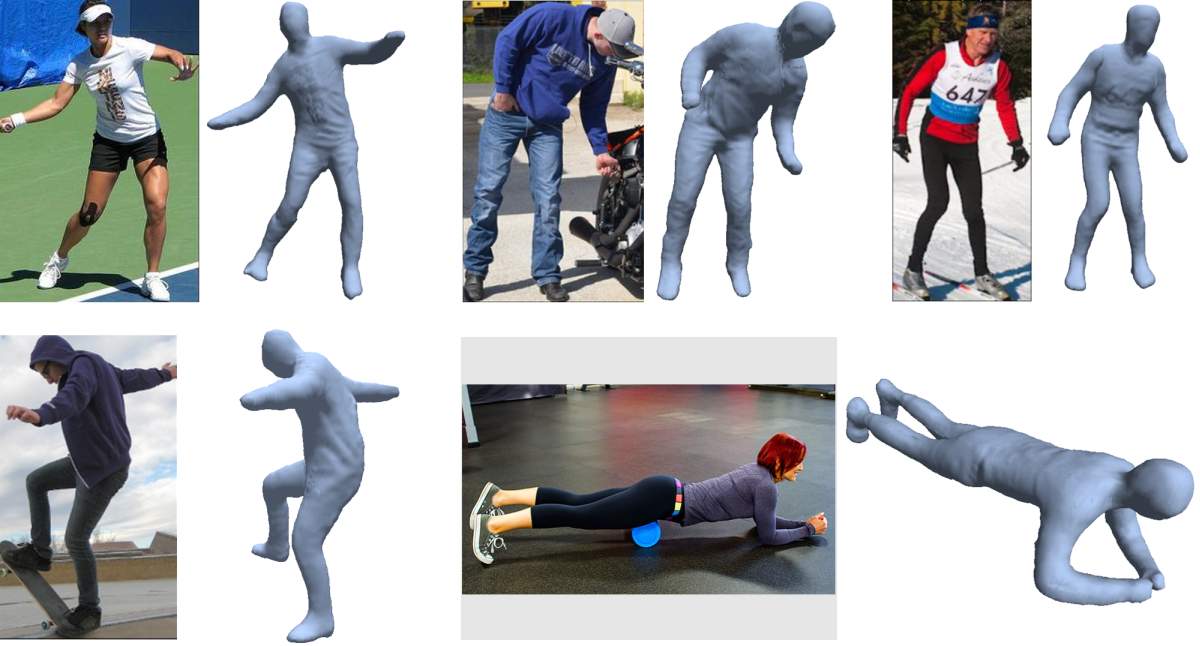}
	\caption{Given only a single RGB image, our method automatically reconstructs the surface geometry of clothed human body. }
	\label{fig:teaser}
\end{figure}

Technically, human reconstruction from a single RGB image is extremely challenging, not only because of the requirement to predict the shape of invisible parts but also due to the need for the geometry recovery for visible surface. Therefore, a method capable of accomplishing such a task should meet two requirements: first, the degrees of freedom of the output space should be constrained to avoid unreasonable artifacts (e.g., broken body parts) in invisible areas; second, the method should be able to efficiently extract geometric information from the input image, such as clothing styles and wrinkles, and fuse them into the 3D space. 

In this paper, we propose DeepHuman, a deep learning-based framework aiming to address these challenges. Specifically, to provide a reasonable initialization for the network and constrain the degrees of freedom of the output space, we propose to leverage parametric body models by generating a 3D semantic volume and a corresponding 2D semantic map as a dense representation after estimating the shape and pose parameters of a parametric body template (e.g., SMPL\cite{SMPL:2015}) for the input image. Note that the requirement of inferring a corresponding SMPL model for an image is not strict; rather, several accurate methods are available for SMPL prediction from a single image\cite{Bogo:ECCV:2016,HMR}. The input image and the semantic volume\&map are fed into an image-guided volume-to-volume translation CNN for surface reconstruction. To accurately recover surface geometry like the hairstyle or cloth contours to the maximum possible extent, we propose a multi-scale volumetric feature transformation so that those different scales of image guidance information can be fused into the 3D volumes. Finally, we introduce a volumetric normal projection layer to further refine and enrich visible surface details according to the input image. This layer is designed to concatenate the volume generation network and the normal refinement network and enables end-to-end training. In summary, we perform 3D human reconstruction in a coarse-to-fine manner by decomposing this task into three subtasks: a) parametric body estimation from the input image, b) surface reconstruction from the image and the estimated body, and c) visible surface detail refinement according to the image. 

The available 3D human dataset \cite{SURREAL} used for network training in BodyNet \cite{BodyNet} is essentially a set of synthesized images textured over SMPL models \cite{SMPL:2015}. No large-scale human 3D dataset with surface geometry under normal clothing is publicly available. To fill in this gap, we present the THuman dataset. We leverage the state-of-the-art DoubleFusion \cite{DoubleFusion} technique for real-time human mesh reconstruction and propose a capture pipeline for fast and efficient capture of outer geometry of human bodies wearing casual clothes with medium-level surface detail and texture. Based on this pipeline, we perform capture and reconstruction of the THuman dataset, which contains about 7000 human meshes with approximately 230 kinds of clothes under randomly sampled poses.

Our network learns from the training corpus synthesized from our THuman dataset. Benefiting from the data diversity of the dataset, the network generalizes well to natural images and provides satisfactory reconstruction given only a single image. We demonstrate improved efficiency and quality compared to current state-of-the-art approaches. We also show the capability and robustness of our method through an extended application on monocular videos.

\section{Related Work}

\textbf{Human Models from Multiview Images.} Previous studies focused on using multiview images for human model reconstruction \cite{Kanade97,StarckCGA07,LiuTVCG2010}. Shape cues like silhouette \cite{IBVH,Loop13}, stereo and shading cues have been integrated in both passive \cite{StarckCGA07,LiuTVCG2010,WuShadingHuman} and active illumination \cite{WaschbuschWCSG05,VlasicPBDPRM09} modes to improve the reconstruction performance. State-of-the-art real-time \cite{Fusion4D,Motion2Fusion} and extremely high-quality \cite{collet2015high} reconstruction results have also been demonstrated with tens or even hundreds of cameras using binocular \cite{UltraStereo} or multiview stereo matching \cite{PMVS} algorithms. To capture detailed motions of multiple interacting characters, more than six hundred cameras have been used to overcome the occlusion challenges \cite{joo2015panoptic,TotalCapture}. However, all these multi-camera systems require complicated environment setups including camera calibration, synchronization and lighting control. 

To reduce the difficulty of system setup, human model reconstruction from extremely sparse camera views has recently been investigated by using CNNs for learning silhouette cues \cite{MinimalCam18} and stereo cues \cite{SparseViewHaoLi18}. These systems require about 4 camera views for a coarse-level surface detail capture. Note also that although temporal deformation systems using lightweight camera setups (usually with about eight cameras) \cite{Vlasic08,Aguiar08,GallSkeleton09} have been developed for dynamic human model reconstruction using skeleton tracking \cite{Vlasic08,LiuPAMI13} or human mesh-based template deformation \cite{Aguiar08}, these systems assume a pre-scanned subject-specific human template as a key model for deformation.   

\textbf{Human Models from Temporal Images.} 
To explore low-cost and convenient human model capture, many studies try to capture a human using only a single RGB or RGBD camera. Since only a single-view camera is needed, methods in this category require the aggregation of multiple temporal frames for full-body model generation. 

For RGBD images, DynamicFusion \cite{newcombe2015dynamic} breaks the static scene assumption and deforms the non-rigid target for TSDF fusion on a canonical static model. Many of the following approaches have tried to improve the robustness by adding color features \cite{Innmann16}, shading constraints \cite{guo2017real} and articulated prior \cite{Yu2017BodyFusion} and dealing with topology changes \cite{Slavcheva17,slavcheva2018cvpr}. The recently appeared DoubleFusion \cite{DoubleFusion} method introduced a human shape prior into the fusion pipeline and achieved state-of-the-art real-time efficiency, robustness, and loop closure performance for efficient human model reconstruction even in cases of fast motions. There are also offline methods for global registration of multiple RGBD images to obtain a full-body model \cite{3DSelfPortrait}.  
To reconstruct a human body using a single-view RGB camera, methods have been proposed for rotating the camera while the target remains as static as possible \cite{ZhuHao17}, or keeping the camera static while the target rotates \cite{VideoAvater,DetailedHumanAvater}. Recently, human performance capture that can reconstruct dynamic human models using only a single RGB camera has been proposed \cite{MonoPerfCap} and sped up to be run in real-time \cite{ReTiCaM}, however, similar to the multicamera scenario \cite{Vlasic08,Aguiar08,GallSkeleton09}, such approaches require a pre-scanned human model obtained using a laser scanner or temporal images-based like \cite{ZhuHao17,VideoAvater,DetailedHumanAvater}.

\textbf{Human Parsing from a Single Image. } Parsing human from a single image has recently been a popular topic in computer vision. The research can be categorized into sparse 2D parsing (2D skeleton estimation) \cite{PoseMachine,DeepCut,OpenPoseCao,Hourglass16}, sparse 3D parsing (3D skeleton estimation) \cite{SimpleYetEffective,Pavlakos2016,Rogez17,Zhou17,Compositional17,LiftingfromtheDeep17,VNect17,3DPoseAdversarial18}, dense 2D parsing \cite{DenseReg,DensePose} and dense 3D parsing (shape and pose estimation). Dense 3D parsing from a single image has attracted substantial interest recently because of the emergence of human statistical models like SCAPE \cite{SCAPE} and SMPL \cite{SMPL:2015}. For example, by fitting the SCAPE or SMPL model to the detected 2D skeleton and other shape cues of an image \cite{Bogo:ECCV:2016,UnitethePeople2017}, or by regressing \cite{HMR,Cipolla2017BMVC,Nips17} the SMPL model using CNNs, the shape and pose parameters can be automatically obtained from a single image. 

Regarding single-view human model reconstruction, there are only two recent works by Varol et al. \cite{BodyNet} and Jackson et al. \cite{ECCVWorkshop18}. In the former study, the 3D human datasets used for the training process are essentially synthesized human imagery textured over SMPL models (lacking geometry details), leading to SMPL-like voxel geometries in their outputs. The latter study shows the ability to output high-quality details, but their training set is highly constrained, leading to difficulty in generalization, e.g., to different human poses.

\begin{figure*}[ht]
	\begin{center}
		\includegraphics[width=1.0\linewidth]{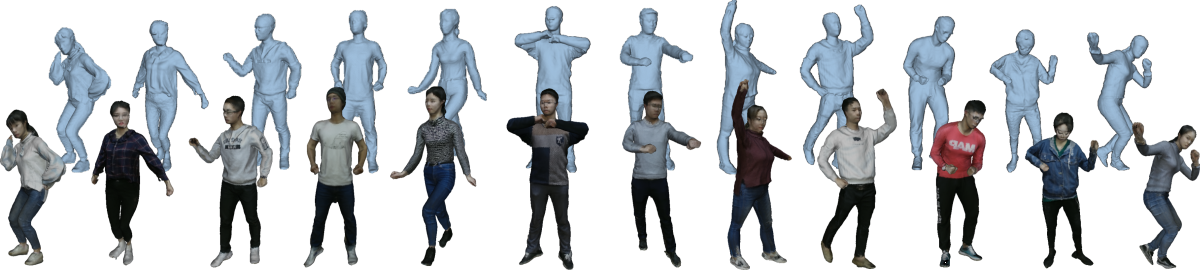}
	\end{center}
	\caption{Example meshes sampled from our dataset. }
	\label{fig:example_data}
\end{figure*}

\textbf{3D Human Body Datasets.}
Most of the available 3D human datasets are used for 3D pose and skeleton detection. Both the HumanEva \cite{HumanEva} and Human3.6M \cite{Human3.6M} datasets contain multiview human video sequences with ground-truth 3D skeleton motion obtained from a marker-based motion capture system. Because of the need to wear markers or special suits, both of these datasets have limited apparel divergence.
MPI-INF-3DHP \cite{MPI-INF-3DHP} dataset enriches the cloth appearance by using a multiview markerless mocap system. However, all the above datasets lack a 3D model of each temporal frame. Recently, with the emergence of the requirement of pose and shape reconstruction from a single image, the synthesized SURREAL \cite{SURREAL} datasets have been created for this task by rendering SMPL models with different shape and pose parameters under different clothing textures. The ``Unite the People" dataset \cite{UnitethePeople2017} provides real-world human images annotated semi-automatic with 3D SMPL models. These two datasets, in contrasts to our dataset, do not contain surface geometry details. 

\section{Overview}
\label{sec:overview}
Given an input image of a person in arbitrary clothes, denoted by $\mathbf{I}$, our method aims to reconstruct his/her full-body 3D surface with plausible geometrical details. Directly reconstructing a 3D surface model of the subject from the given image is very challenging because of depth ambiguities, body self-occlusions and high degree of freedom of the output space. For this reason we perform 3D human reconstruction in a coarse-to-fine manner. Our method starts with body estimation, then steps into surface reconstruction and finally recovers the details on the frontal areas of the surface. 

To estimate a body from $\mathbf{I}$, we exploit the state-of-the-art methods, HMR\cite{HMR} and Simplify\cite{Bogo:ECCV:2016}, both of which are able to infer the shape and pose parameters of SMPL\cite{SMPL:2015} from a single image. We found that they have complementary characteristics: HMR's prediction are always plausible but not well-aligned with the color image, while Simplify aligns the SMPL model with detected keypoint very well but relies on initialization to output plausible results. To obtain an accurate SMPL estimation for the image, we first use HMR to obtain an initial estimation and then improve its accuracy using Simplify. 

The estimated body shape and pose parameters determine a polygon mesh representation of the body through linear shape blending and pose skinning\cite{SMPL:2015}. However, it is hard to feed the polygon mesh representation into a deep neural network. Therefore, inspired by ``Vitruvian Manifold" \cite{VitruvianManifold}, we introduce a dense semantic representation generated from SMPL. Specifically, we pre-define a \textit{semantic code} (a 3-dimensional vector) for each vertex on SMPL according to its spatial coordinate at rest pose. Given a SMPL model corresponding to a human image, we render the semantic code onto the image plane to obtain a semantic map $\mathbf{M}_s$, and generate a semantic volume $\mathbf{V}_s$ by first voxelizing the SMPL model into voxel grid and then propagating the semantic codes into the occupied voxels. Our dense semantic representation has three advantages: it encodes information about both the shape and the pose of the body, provides clues about the corresponding relationship between 3D voxels and 2D image pixels, and is easy to be incorporated into neural networks. 

For the surface geometry reconstruction, we adopt an occupancy volume to represent the surface\cite{BodyNet}. Specifically, we define a 3D occupancy voxel grid $\mathbf{V}_o$, where the values of voxels inside the surface are set to $1$ and others are set to $0$. All occupancy volumes have a fixed resolution of $128\times192\times128$, where the resolution of the y-axis is set to a greater value because we observed that a 3D human model usually has a major axis. To reconstruct $\mathbf{V}_o$ from $\mathbf{V}_s$ with the assistance of $\mathbf{I}$ and $\mathbf{M}_s$, we propose an image-guided volume-to-volume translation network(Sec.\ref{sec:network_arch}), in which we use multi-scale volumetric feature transformation(Sec.\ref{sec:vft}) to fuse 2D image guidance information into a 3D volume. Accordingly, the network will take advantage of knowledge from both the 2D image and the 3D volume. 

Due to resolution limitations, a voxel grid always fails to capture or recover fine details like wrinkles on the clothes. To further enrich the geometrical details on the visible part of the surface, we propose to directly project a 2D normal map $\mathbf{N}$ from $\mathbf{V}_o$ (Sec.\ref{sec:normal_proj}) and refine it with a U-net(Sec.\ref{sec:network_arch}). In other words, we encode the geometrical details of the frontal surface using 2D normal maps, and consequently lower the memory requirement.

To train the network with supervision, we contribute THuman, a real-world 3D human model dataset (Sec.\ref{sec:dataset}). We synthesize training corpus from the dataset. Once the network is trained, it can be used to predict an occupancy volume and a frontal normal map given an RGB image of a person. We obtain the final polygon mesh model for the subject by firstly extracting a triangular polygon mesh from the occupancy volume using Marching Cube algorithm and then refining the mesh according to the frontal normal map using the method in \cite{Nehab2005Efficiently}.

\section{Approach}
\label{sec:network}
\begin{figure*}[h]
	\begin{center}
		\includegraphics[width=0.95\linewidth]{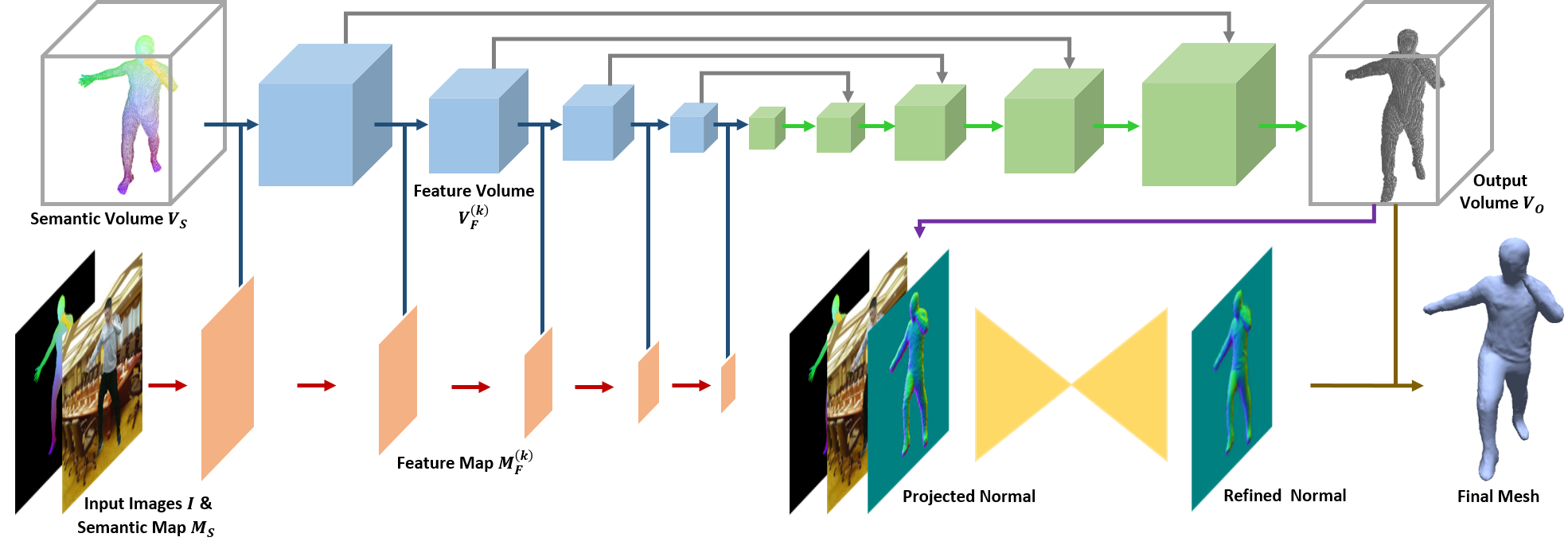}
		\includegraphics[width=0.8\linewidth]{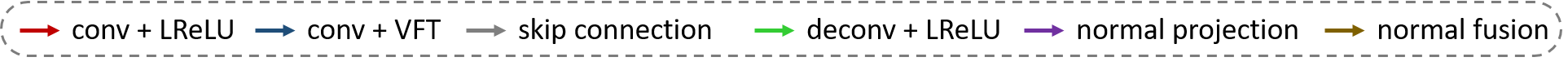}
	\end{center}
	\caption{Network architecture. Our network is mainly composed of an image feature encoder (orange), a volume-to-volume translation network (blue \& green) and a normal refinement network (yellow). }
	\label{fig:net}
\end{figure*}

\subsection{Network Architecture}
\label{sec:network_arch}
Our network consists of 3 components, namely an image feature encoder $\mathcal{G}$, a volume-to-volume (vol2vol) translation network $\mathcal{H}$ and a normal refinement network $\mathcal{R}$, as shown in Fig.\ref{fig:net}. The image feature encoder $\mathcal{G}$ aims to extract multi-scale 2D feature maps $\mathbf{M}_f^{(k)} (k=1, \ldots, K)$ from the combination of $\mathbf{I}$ and $\mathbf{M}_s$. The vol2vol network is a volumetric U-Net~\cite{Yang2018Dense}, which takes $\mathbf{V}_s$ and $\mathbf{M}_f^{(k)} (k=1, \ldots, K)$ as input, and outputs an occupancy volume $\mathbf{V}_o$ representing the surface. Our vol2vol network $\mathcal{H}$ fuses multi-scale semantic features $\mathbf{M}_f^{(k)} (k=1, \ldots, K)$ into its encoder through a \textit{multi-scale volumetric feature transformer}. After generating $\mathbf{V}_o$, a normal refinement U-Net~\cite{UNet2015} $\mathcal{R}$ further refines the normal map $\mathbf{N}$ after calculating it directly from $\mathbf{V}_o$ through a \textit{volume-to-normal projection layer}. All operations in the network are differentiable, and therefore, it can be trained or fine-tuned in an end-to-end manner. 

\subsubsection{Multi-scale Volumetric Feature Transformer} 
\label{sec:vft}
\begin{figure}
	\centering
	\includegraphics[width=1.0\linewidth]{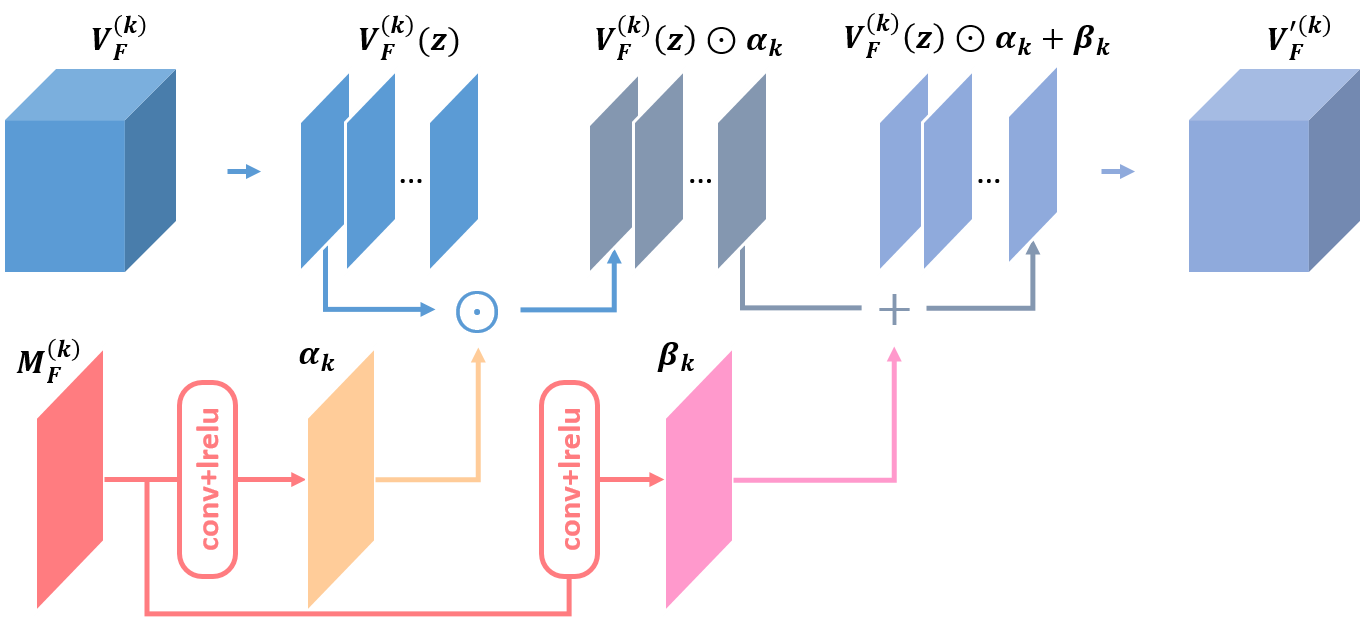}
	\caption{Illustration of volumetric feature transformation at level $k$. }
	\label{fig:vft}
\end{figure}
In this work, we extend the Spatial Feature Transformer (SFT) layer \cite{Wang2018Recovering} to handle 2D-3D data pairs in the multi-scale feature pyramid, and propose multi-scale Volumetric Feature Transformer (VFT). SFT was first used in \cite{Wang2018Recovering} to perform image super-resolution conditioned on semantic categorical priors to avoid the regression-to-the-mean problem. A SFT layer learns to output a modulation parameter pair $(\alpha, \beta)$ based on the input priors. Then transformation on the feature map $\mathbf{F}$ is carried out as: 
\begin{equation}
\label{eqn:sft}
\mathcal{SFT}(\mathbf{F}) = \alpha\odot\mathbf{F} + \beta
\end{equation}
where $\odot$ is Hadamard product.

In our network, at each level $k$, a feature volume $\mathbf{V}_f^{(k)}$ (blue cubes in Fig.\ref{fig:net}) and a feature map $\mathbf{M}_f^{(k)}$ (orange squares in Fig.\ref{fig:net}) are provided by previous encoding layers.  Similar to \cite{Wang2018Recovering}, we first map the feature map $\mathbf{M}_f^{(k)}$ to modulation parameters $(\alpha_k, \beta_k)$ through convolution+activation layers (see the second row of Fig.\ref{fig:vft}). Note that the operation in Eqn.(\ref{eqn:sft}) cannot be applied directly on $\mathbf{V}_f^{(k)}$ and $\mathbf{M}_f^{(k)}$ because of dimension inconsistency ($\mathbf{V}_f^{(k)}$ has a z-axis while $(\alpha_k, \beta_k)$ doesn't.) Therefore, we slice the feature volume along the $z$-axis into a series of feature slices, each of which has a thickness of $1$ along the $z$-axis. Then we apply the same element-wise affine transformation to each feature $z$-slice independently:
\begin{equation}
\mathcal{VFT}\left(\mathbf{V}_f^{(k)}(z_i)\right) = \alpha_k\odot\mathbf{V}_f^{(k)}(z_i) + \beta_k 
\end{equation}
where $\mathbf{V}_f^{(k)}(z_i)$ is the feature slice on plane $z=z_i, z_i=1,2,\ldots,Z$ and $Z$ is the maximal z-axis coordinate. The output of a VFT layer is the re-combination of transformed feature slices. The operations applied by VFT layers are illustrated in Fig.\ref{fig:vft}.

The superiority of VFT is three-fold. Firstly, compared to converting feature volumes/maps into a latent codes and concatenating them at the network bottleneck, it preserves the shape primitiveness of feature maps and thus encodes more local information. Second, it is efficient. Using VFT, feature fusion can be achieved in one single pass of affine transformation, without requiring extra convolutions, full connection or other operations. Third, it is flexible. VFT can be performed on either the original image/volume or downsampled feature maps/volumes, which makes it possible to fuse different scales of features, enabling much deeper feature transfer.

In order to integrate image features to the maximum possible extent, we perform volumetric feature transformation on the multi-scale feature pyramid; see the blue arrows/lines in Fig.\ref{fig:net} for illustration. We only perform VFT in the encoder part of our vol2vol network; however, the transformation information can be propagated to the decoder through skip-connections. As discussed in Sec.\ref{sec:evaluation}, the multi-scale feature transformation helps preserve geometry details compared to directly concatenating latent variables at the network bottleneck.

\subsubsection{Volume-to-normal Projection Layer}
\label{sec:normal_proj}
\begin{figure}
	\centering
	\includegraphics[width=1.0\linewidth]{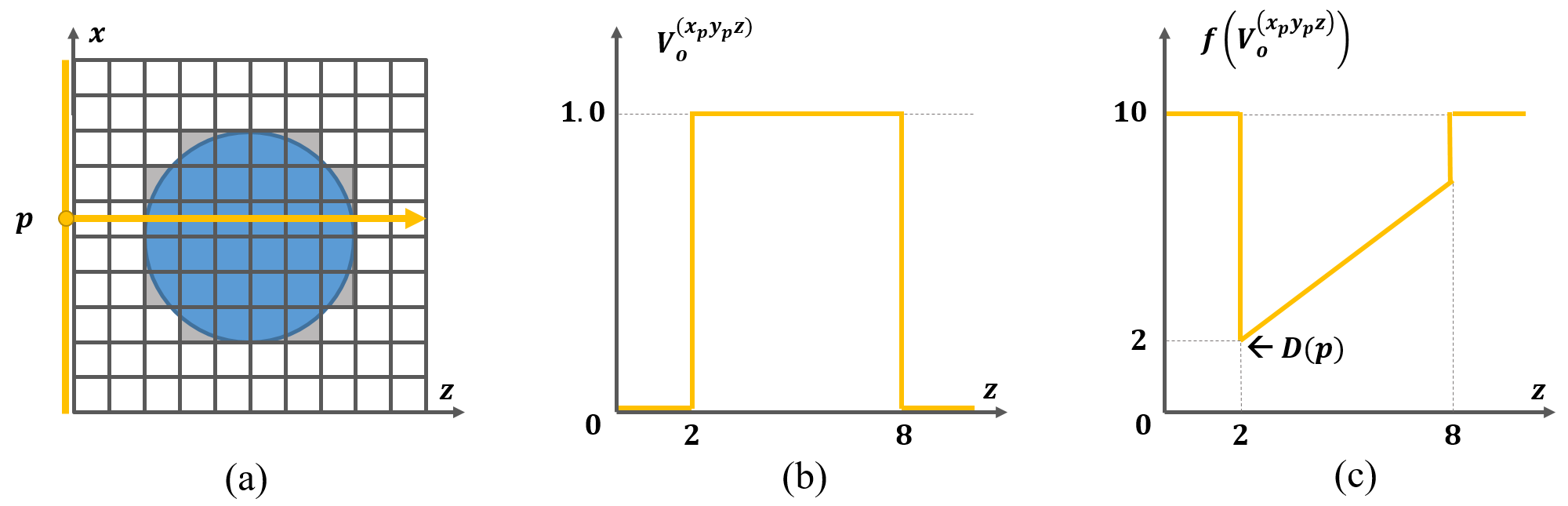}
	\caption{Illustration of differentiable depth projection. }
	\label{fig:depth_projector} 
\end{figure}

Our goal is to recover geometrical details (e.g. wrinkles and cloth boundary) on the visible surface of the human model. However, a volume-based representation is unable to recover such fine-grain details due to resolution limitations. Thus, we encode the frontal geometrical details on 2D normal maps, which can be directly calculated from the occupancy volume using our differentiable volume-to-normal projection layer. The layer first projects a depth map directly from the occupancy volume, transforms the depth map into a vertex map, and then calculates the normal maps through a series of mathematical operations. 

Fig.\ref{fig:depth_projector} is a 2D illustration explaining how the layer projects depth maps. In Fig.\ref{fig:depth_projector}(a), the blue circle is the model we aim to reconstruct, and the voxels occupied by the circle are marked in grey. Consider the pixel $p = (x_p, y_p)$ on the image plane as an example. To calculate depth value $\mathbf{D}(p)$ of $p$ according to $\mathbf{V}_o$, a straightforward method is to consider a ray along the $z$-axis and record the occupancy status of all voxels along that ray (Fig.\ref{fig:depth_projector}(b)). Afterwards, we can determine $\mathbf{D}(p)$ by finding the nearest occupied voxel. Formally, $\mathbf{D}(p)$ is obtained according to
\begin{equation}
\mathbf{D}(p) = \inf \left\{ z|\mathbf{V}_o^{(x_py_pz)} = 1 \right\}
\end{equation}
where $\mathbf{V}_o^{(x_py_pz)}$ denotes the value of the voxel at coordinate $(x_p, y_p, z)$. Although this method is straightforward, it is difficult to incorporate the operation, $\inf\{\cdot\}$ into neural networks due to the complexity of differentiating through it. Therefore, we transform the occupancy volume to a \textit{depth volume} $\mathbf{V}_d$ by applying a transformation $f$: 
\begin{equation}
\mathbf{V}_d^{(xyz)} = f(\mathbf{V}_o^{(xyz)}) = M(1-\mathbf{V}_o^{(xyz)}) + z\mathbf{V}_o^{(xyz)}
\end{equation}
where $M$ is a sufficiently large constant. Then $\mathbf{D}(p)$ can be computed as:
\begin{equation}
\mathbf{D}(p) = \min_z f(\mathbf{V}_d^{(x_py_pz)}),  
\end{equation}
as illustrated in Fig.\ref{fig:depth_projector}(c). 

After depth projection, we transform the depth map to a vertex map $\mathbf{M}_v$ by assigning $x$ and $y$ coordinates to depth pixels according to their positions on the images. Then Sobel operators are used to calculate the directional derivative of the vertex map along both the $x$ and $y$ directions: $\mathbf{G}_x = S_x * \mathbf{M}_v, \mathbf{G}_y = S_y * \mathbf{M}_v$, where $S_x$ and $S_y$ are Sobel operators. The normal at pixel $p = (x_p, y_p)$ can be calculated as:
\begin{equation}
\mathbf{N}^{(x_py_p)} = \mathbf{G}_x(p) \times \mathbf{G}_y(p), 
\end{equation}
where $\times$ denotes cross product. Finally, $\mathbf{N}$ is up-sampled by a factor of 2 and further refined by a U-Net.  

\subsection{Loss Functions}
\label{sec:loss}
Our loss functions used to train the network parameters consist of reconstruction errors for the 3D occupancy field and 2D silhouette, as well as the reconstruction loss for normal map refinement. We use extended Binary Cross-Entropy (BCE) loss for the reconstruction of occupancy volume~\cite{jackson2017vrn}:
\begin{equation}
\label{eqn:L_V}
\begin{split}
\mathcal{L}_{V} = -\frac{1}{\left|\mathbf{\hat{V}}_o\right|}\sum_{x,y,z}&\gamma\mathbf{\hat{V}}_o^{(xyz)}\log\mathbf{{V}}_o^{(xyz)} +\\ &(1-\gamma)\left(1-\mathbf{\hat{V}}_o^{(xyz)}\right)\log\left(1-\mathbf{{V}}_o^{(xyz)}\right)
\end{split}
\end{equation}
where $\mathbf{\hat{V}}_o$ is the ground-truth occupancy volume corresponding to $\mathbf{V}_o$, $\mathbf{{V}}_o^{(xyz)}$ and $\mathbf{\hat{V}}_o^{(xyz)}$ are voxels in the respective volumes at coordinate $(x, y, z)$, and $\gamma$ is a weight used to balance the loss contributions of occupied and unoccupied voxels. Similar to \cite{BodyNet}, we use a multi-view re-projection loss on the silhouette as additional regularization:
\begin{equation}
\begin{split}
\mathcal{L}_{FS} = -\frac{1}{\left|\mathbf{\hat{S}}_{fv} \right|} \sum_{x,y} &\mathbf{S}_{fv}^{(xy)}\log\mathbf{{S}}_{fv}^{(xy)} +\\
&\left(1-\mathbf{\hat{S}}_{fv}^{(xy)}\right)\log\left(1-\mathbf{{S}}_{fv}^{(xy)}\right)
\end{split}
\end{equation}
where $\mathcal{L}_{FS}$ denotes the front-view silhouette re-projection loss, $\mathbf{{S}}_{fv}$ is the silhouette re-projection of $\mathbf{{V}}_o$, $\mathbf{\hat{S}}_{fv}$ is the corresponding ground-truth silhouette, and $\mathbf{{S}}_{fv}^{(xy)}$ and $\mathbf{\hat{S}}_{fv}^{(xy)}$ denote their respective pixel values at coordinate $(x, y)$. Assuming a weak-perspective camera, we can easily obtain $\mathbf{S}_{fv}^{(xy)}$ through orthogonal projection~\cite{BodyNet}: $\mathbf{{S}}_{fv}^{(xy)} = \max_z \mathbf{{V}}_o^{(xyz)}$. The side-view re-projection loss $\mathcal{L}_{SS}$ is defined similarly. 

For normal map refinement, we use the cosine distance to measure the difference between predicted normal maps and the corresponding ground truth: 
\begin{equation}
\mathcal{L}_{N} = \frac{1}{\left|\mathbf{\hat{N}} \right|} \sum_{x,y} 1-\frac{<\mathbf{N}^{(xy)}, \mathbf{\hat{N}}^{(xy)}>}{|\mathbf{N}^{(xy)}|\cdot|\mathbf{\hat{N}}^{(xy)}|}
\end{equation}
where $\mathbf{N}^{(xy)}$ is the refined normal map produced by the normal refiner, $\mathbf{\hat{N}}^{(xy)}$ is the ground-truth map, and similarly $\mathbf{N}^{(xy)}$ and $\mathbf{\hat{N}}^{(xy)}$ denote their respective pixel values at coordinate $(x, y)$. 

Therefore, the combined loss is
\begin{equation}
\mathcal{L} = \mathcal{L}_{V} + \lambda_{FS}\mathcal{L}_{FS} + \lambda_{SS}\mathcal{L}_{SS} + \lambda_{N}\mathcal{L}_{N},
\end{equation}
where $\lambda_{\bullet}$s are scalar weights of the loss terms.

\subsection{Implementation details}
\label{sec:implementation-detail}
\begin{table}[]
	\footnotesize
	\caption{Network Architecture Details. }
	\label{tab:net_arch}
	\centering
	\begin{threeparttable}
		\begin{tabular}{ccccc}
			\hline
			Net & Layer & Kernel & Stride & Output\\
			\hline
			\hline
			\multirow{5}{*}{$\mathcal{G}$} & conv+lrelu & 4 & 2 & $96\times64\times8$\\
			~  & conv+lrelu & 4 & 2 & $48\times32\times16$\\
			~  & conv+lrelu & 4 & 2 & $24\times16\times32$\\
			~  & conv+lrelu & 4 & 2 & $12\times8\times64$\\
			~  & conv+lrelu & 4 & 2 & $6\times4\times128$\\
			\hline
			\multirow{11}{*}{$\mathcal{H}$} & conv+lrelu & 4 & 2 & $64\times96\times64\times8$\\
			~  & conv+lrelu & 4 & 2 & $32\times48\times32\times16$\\
			~  & conv+lrelu & 4 & 2 & $16\times24\times16\times32$\\
			~  & conv+lrelu & 4 & 2 & $8\times12\times8\times64$\\
			~  & conv+lrelu & 4 & 2 & $4\times6\times4\times128$\\
			~  & transconv+lrelu & 4 & 2 & $8\times12\times8\times64$\\
			~  & transconv+lrelu & 4 & 2 & $16\times24\times16\times32$\\
			~  & transconv+lrelu & 4 & 2 & $32\times48\times32\times16$\\
			~  & transconv+lrelu & 4 & 2 & $64\times96\times64\times8$\\
			~  & transconv+lrelu & 4 & 2 & $128\times192\times128\times4$\\
			~  & conv+sigmoid & 3 & 1 & $128\times192\times128\times1$\\
			\hline
			\multirow{9}{*}{$\mathcal{R}$} & conv+lrelu & 4 & 2 & $192\times128\times16$ \\
			~ & conv+lrelu & 4 & 2 & $96\times64\times32$ \\
			~ & conv+lrelu & 4 & 2 & $48\times32\times32$ \\
			~ & conv+lrelu & 4 & 2 & $24\times16\times32$ \\
			~ & conv+lrelu & 4 & 2 & $12\times8\times32$ \\
			~ & transconv+lrelu & 4 & 2 & $24\times16\times32$ \\
			~ & transconv+lrelu & 4 & 2 & $48\times32\times32$ \\
			~ & transconv+lrelu & 4 & 2 & $96\times64\times32$ \\
			~ & transconv+lrelu & 4 & 2 & $192\times128\times16$ \\
			~ & transconv+lrelu & 4 & 2 & $384\times256\times8$ \\
			~ & conv+tanh & 3 & 1 & $384\times256\times3$ \\
			\hline
		\end{tabular}
		
		\begin{tablenotes}
			\item[*] The term ``conv" is convolution for short, ``transconv" is transposed convolution and ``lrelu" is Leakly ReLU. 
		\end{tablenotes}
		
	\end{threeparttable}
\end{table}

The volume-to-volume network $\mathcal{H}$ takes as input a semantic volume with $128\times192\times128$ resolution, and outputs an occupancy with the same shape. The image encoder $\mathcal{G}$ concatenates as input the given RGB image and the corresponding semantic map, both of which have a resolution of $192\times128$.
Our normal refinement U-Net $\mathcal{R}$ takes as input the concatenation of the RGB image, semantic map and upsampled normal projection result, and the input/output resolution of $\mathcal{R}$ is $384\times256$. The architecture details are shown in Tab.\ref{tab:net_arch}. 

During network training, the parameters are set to $\lambda_{FS}=\lambda_{SS}=0.1, \lambda_{N}=0.01, \gamma=0.7$. We exploit a two-stage training procedure: first pre-train the vol2vol network and the normal refinement network, and then fine-tune them jointly with the combined loss. We used Adam~\cite{adam2014} with default parameters as the optimizer. The learning rate is fixed to 2e-4 during the whole training procedure, and the batch size is set to be 4.

\section{THuman: 3D Real-world Human Dataset}
\label{sec:dataset}
Collecting rich 3D human surface model with texture containing casual clothing, various human body shapes and natural poses has been a time-consuming and laborious task as it always relies on either expensive laser scanners or sophisticated multiview systems in a controlled environment. Fortunately, this task becomes easier with the recently introduced real-time human performance capture system using only a single depth camera \cite{DoubleFusion,Zheng2018HybridFusion}. In this section we present the method to capture a 3D real-world human dataset called  ``THuman dataset" for later supervised deep learning on single-image human reconstruction.

Our capture system is based on the single-view RGB-D DoubleFusion~\cite{DoubleFusion} technique. DoubleFusion utilizes a double-layer representation and incorporates a motion prior derived from the SMPL \cite{SMPL:2015}. It simultanuously solves skeleton motions and non-rigid deformation according to the depth observation at the current frame. After getting the motion field, depth pixels in the current frame are fused into a reference volume as described in \cite{newcombe2015dynamic}. As the observed surface is gradually fused and deformed, the shape and pose parameters of the body layer are also gradually optimized through volumetric shape-pose optimization. In this way the two layers can benefit from each other, leading to robust tracking and accurate reconstruction. 

The available DoubleFusion technique performs only robust fusion of detailed surface geometries. To obtain full-body texture, we can directly perform color or albedo fusion in a similar way to depth fusion. However, the fused texture blurs when fast body motion occurs. Thus we develop a two-stage capture procedure. In the first stage, the subject actors are required to rotate slowly and perform some basic surface completion motions to obtain a surface geometry that is as complete as possible and clear texture recovery of the surface as well. After that, in the second stage, we disable geometry fusion and texture update, but still perform the non-rigid surface registration based on the input depth information. In this way, we still capture non-rigid motion details of the subject's surface.

In order to obtain human mesh data under natural but diverse poses, our system presents to the subject a reference pose randomly sampled from MOSH\cite{Loper:SIGASIA:2014} dataset every 6 seconds and requires the performer to imitate the reference pose in the second stage. Note that the 6-second interval is usually long enough for subjects to recognize the presented pose and prepare for imitation. At the end of every 6-second interval, the system automatically saves the RGBD image, the 3D surface mesh and its corresponding SMPL model in the current live pose. After data capture, we post-process the raw meshes through hole filling \cite{Kazhdan2013Screened}, remeshing \cite{Jakob2015Instant} and isolated artifact removal. 

After approximately 70 hours of data capture using only one capture setup, we achieve capturing and reconstruction of 230 subject characters, with each character corresponding to about 30 poses. This data leads to 7000 data items in our THuman dataset; some examples are shown in Fig.~\ref{fig:example_data}. Each item contains a textured surface mesh, a RGBD image from the Kinect sensor, and an accompanying well-aligned SMPL model. 

In this work, we only use the textured surface mesh and the accompanied SMPL model to generate training data. The training corpus are synthesized in the following steps: for each model in our dataset, we first render 4 color images from 4 random viewpoints using a method similar to \cite{SURREAL}; after that, we generate the corresponding semantic maps and volumes, occupancy volumes as well as normal maps. By enumerating all the models in our dataset, we finally synthesize $\sim${28}K images for network training. 

\section{Experiments}

\subsection{Results}
\label{sec:results}
\begin{figure*}
	\centering
	\includegraphics[width=1.0\linewidth]{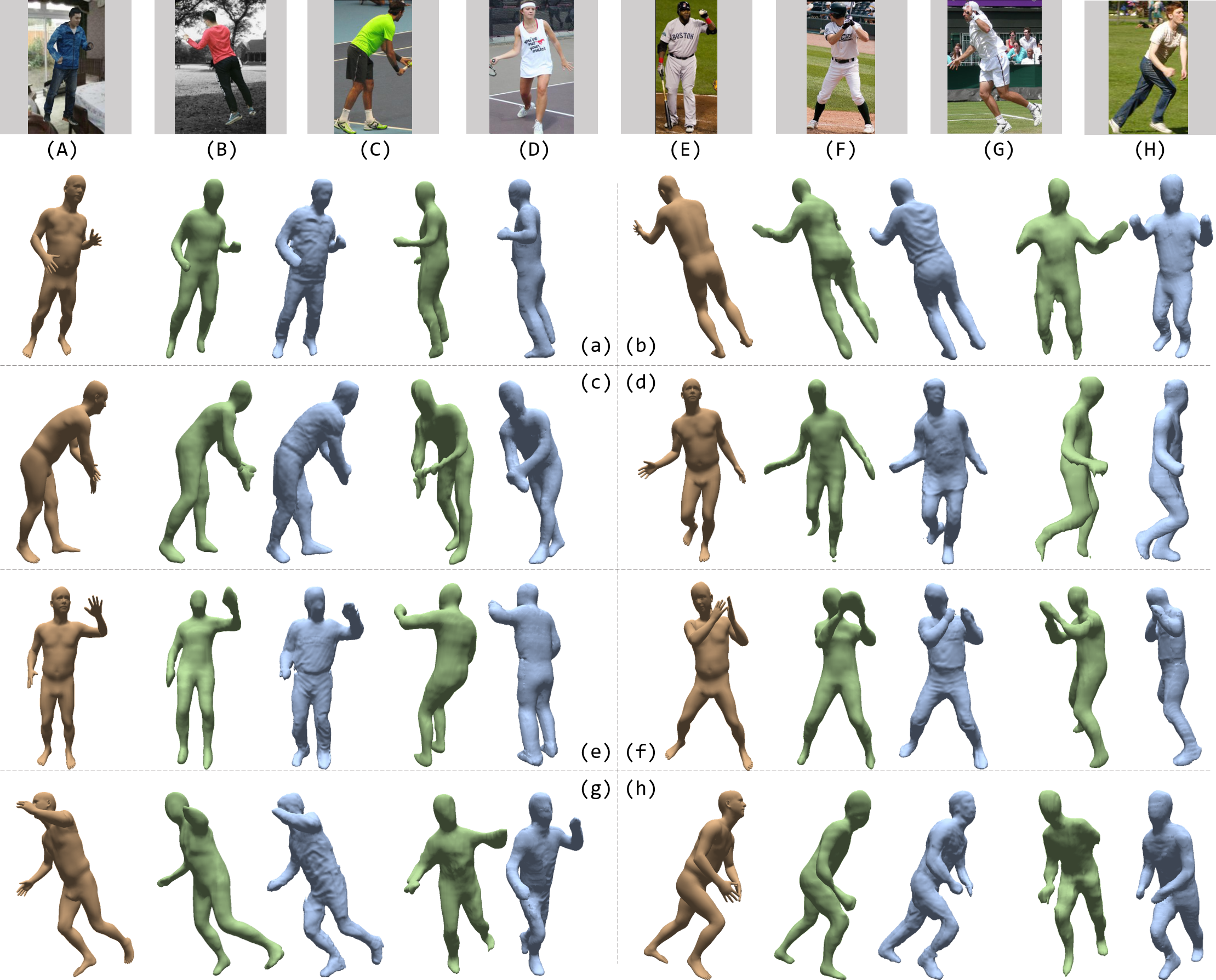}
	\caption{Reconstruction results on synthesis images and natural images. The input images are presented in the first row, while last four rows show the results of HMR\cite{HMR} (in orange), BodyNet\cite{BodyNet} (in green) and our method (in blue). For BodyNet and our method we render the results from two views, i.e., a front view and a side view. }
	\label{fig:results_large}
\end{figure*}
\begin{figure*}
	\centering
	\includegraphics[width=1.0\linewidth]{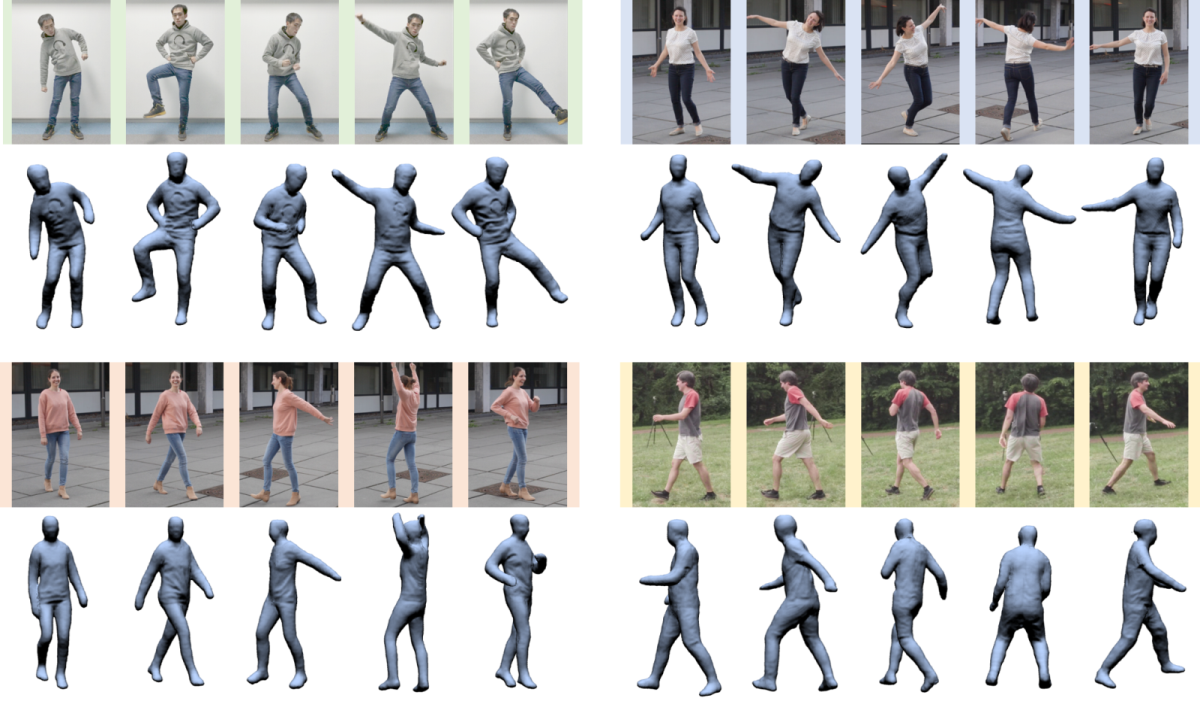}
	\caption{3D reconstruction from a monocular video using our method. }
	\label{fig:vid_app}
\end{figure*}

We demonstrate our approach with a range of photos of people in Fig.\ref{fig:results_large}. In this figure, image (A) and (B) are testing images synthesized from our THUman dataset, while image (C)-(H) are natural images sampled from the LIP dataset\cite{Gong_2017_LIP}. Note that the subjects in image (A) and (B) do not appear in our training data. As shown in Fig.\ref{fig:results_large} our approach is able to reconstruct both the 3D human models and surface details like cloth wrinkles (see Fig.\ref{fig:results_large}(a,b,c)). From Fig.\ref{fig:results_large}(d,e,f) we can also see that our method can not only reconstruct the 3D meshes according to input images and detected SMPL models, but also recover discontinuities on the surface like a belt and the hem of a dress. In Fig.\ref{fig:vid_app} we show an extended application on 3D human performance capture from a single-view RGB video. It should be noted that the reconstruction results are generated by applying our method on each the video frame independently, without any temporal smoothness involved. The results demonstrate the ability of our method to tackle various human poses and its robust performance. Please see the supplemental video for more details.

\subsection{Comparison}
\subsubsection{Competing Approaches}
We compare our method against two two state-of-the-art deep learning based approaches for single view 3D human reconstruction. To eliminate the effect of dataset bias, we fine-tuned the pre-trained model of both network with the same training data as we use to train our network. 

(1) \textbf{HMR}. In \cite{HMR}, Kanazawa et al. proposed a neural network to directly regress the shape and pose parameter of SMPL from an RGB image. The output of HMR is a 75-D vector, which can be used to generate a triangular mesh of SMPL through linear shape blending and pose skinning\cite{SMPL:2015}. It is the state-of-the-art among available methods for single-view pose and shape estimation\cite{HMR,Bogo:ECCV:2016,GuanICCV2009,Cipolla2017BMVC,Nips17}. We fine-tuned the pretrained HMR model using the color images and the corresponding ground-truth shape/pose parameters in our synthesis training data. 

(2) \textbf{BodyNet}. In \cite{BodyNet}, Varol et al. proposed a neural network for direct inference of volumetric body shape from a single image. The output of BodyNet is a $128\times128\times128$ occupancy volume with similar definition in Sec.\ref{sec:overview}. BodyNet is the most related work to this paper. We fine-tuned the whole network of BodyNet using the color images and the ground-truth occupancy volumes in our training set.  

\subsubsection{Comparison Results}
We compare against HMR and BodyNet both qualitatively and quantitatively. 

(1) For \textbf{qualitative comparison}, we feed all the networks with the same images and convert the network output into a triangular mesh. The results are rendered in Fig.\ref{fig:results_large}. As shown in the figure, our method is able to achieve much more detailed reconstruction than HMR and BodyNet (See Fig.\ref{fig:results_large}(a$\sim$f)). In addition, our method has higher robustness than BodyNet when some body parts are occluded  (See Fig.\ref{fig:results_large}(b,g,h)). 

(2) The \textbf{quantitative comparison} is conducted on the testing set of our synthetic data. We convert the output of HMR to occupancy volumes with a resolution of $128\times192\times128$. We also upsampled the output of BodyNet by a factor of $1.5$ using trilinear interpolation, and then crop the volume to make it have the same resolution. After that we use the mean Intersection-over-Union (IoU score) between predicted 3D volumes and their ground-truth as the comparison metric. It should be noted that the predicted volume and the ground-truth may be unaligned along the $z$-axis because of depth ambiguities. Therefore, we shift the predicted volume along $z$-axis to search for the best alignment (i.e., to maximize IoU score between the ground-truth volumes and the predict ones), and regard the maximum IoU score as the final score. The results are presented in Tab.\ref{tab:quant_comp}. As shown by the numerical results, our method achieves the most accurate reconstruction among all the approaches. BodyNet occasionally produces broken bodies and consequently gets the lowest score. 

\begin{table}[h]
	\caption{Quantitative comparison using 3D IOU score. }
	\centering
	\begin{tabular}{c|c}
		\hline
		Method & Averaged 3D IOU\\
		\hline
		HMR     & 41.4\% \\
		BodyNet & 38.7\% \\
		Ours    & 45.7\% \\
		\hline
	\end{tabular}
	\label{tab:quant_comp}
\end{table}

\subsection{Ablation Study}
\label{sec:evaluation}

\subsubsection{Semantic Volume/Map Representation}
\begin{table}[h]
	\caption{Numerical evaluation of semantic volume/map representation. }
	\centering
	\begin{tabular}{l|c}
		\hline
		Representation & IOU score (\%)  \\
		\hline
		Joints Heat Map/Volume & 74.16 \\
		Semantic Map/Volume & 79.14  \\
		\hline
	\end{tabular}
	\label{tab:reprez_eval}
\end{table}
\begin{figure}[h]
	\centering
	\includegraphics[width=1.0\linewidth]{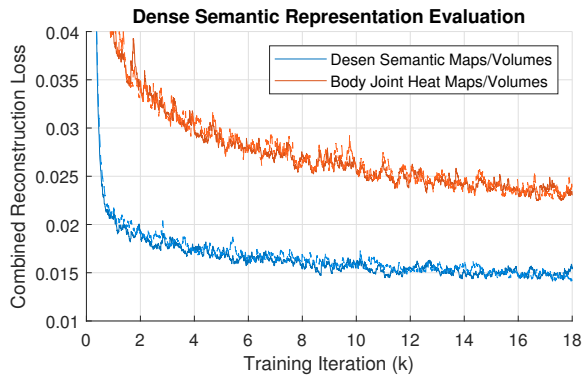}
	\caption{Evaluation of semantic volume/map representation. We evaluate two different inputs for the image-guided vol2vol network: dense semantifc maps/volumes and body joints heat maps/volumes. We show the combined reconstruction losses ($\mathcal{L}_{V} + \lambda_{FS}\mathcal{L}_{FS} + \lambda_{SS}\mathcal{L}_{SS}$) for different inputs. Solid lines show training error, while dashed lines show validation error (they almost overlap with each other). }
	\label{fig:reprez_eval_loss}
\end{figure}

\noindent\textbf{Baseline. } An alternative representation to our semantic volume/map is body joint heat volumes/maps that are used in BodyNet\cite{BodyNet}. A joint heat map is a multi-channel 2D image where in each channel a Gaussian with fixed variance is centered at the image location of the corresponding joint. By extending the notion of 2D heat maps to 3D, we can also define the heat volumes for body joints. In order to evaluate our semantic volume/map representation, we implement a baseline network that takes body joints' heat maps and heat volumes as input and has the identical structure to the network presented in Sec.\ref{sec:network}. In this experiment we generate input semantic volumes/maps and joint heat volumes/maps from the ground-truth SMPL model to eliminate the impact of inaccurate SMPL estimation. 

\noindent\textbf{Results. } Fig.\ref{fig:reprez_eval_loss} shows the evaluation results. The figure shows that compared to sparse joints, a network taking dense semantic maps/volumes is able to learn to reconstruct the 3D model more accurately. In Tab.\ref{tab:reprez_eval}, we also test these two methods on the testing portion of our synthetic dataset and measure the reconstruction error by calculating the IoU score between the network output and the ground-truth volume. The numerical results also show that taking dense semantic maps/volumes as input helps the network achieve higher reconstruction accuracy. We think that it may be because our semantic volume/map representation encodes information about the body shape and pose jointly and provides a good initialization for the volumetric reconstruction network.

\subsubsection{Multi-scale Volumetric Feature Transformation}
\begin{figure*}
	\centering
	\includegraphics[width=1.0\linewidth]{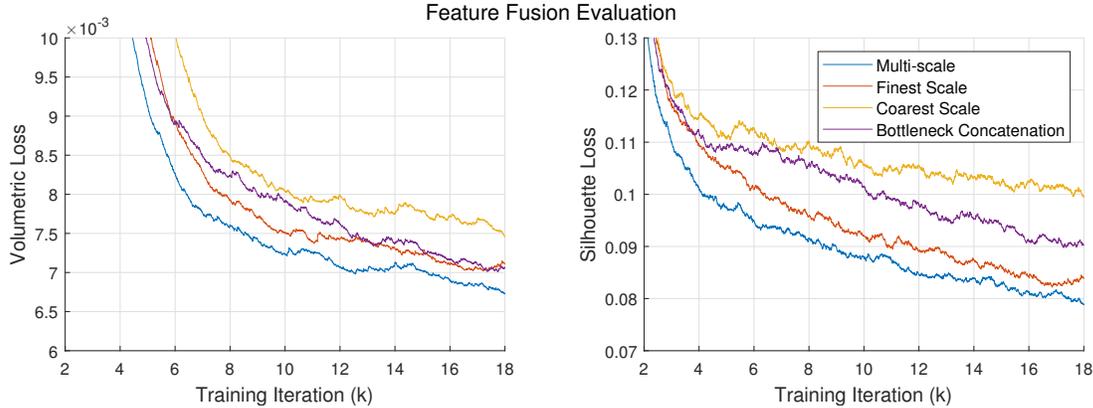}
	\caption{Evaluation of multi-scale volumetric feature transformation (VFT). We evaluate several ways to fuse 2D features into 3D volumes, and show the volumetric loss ($\mathcal{L}_{V}$) and the silhouette loss ($\mathcal{L}_{FS} + \mathcal{L}_{SS}$) in the figure. For clarity we do not show the validation loss. }
	\label{fig:vft_eval_loss}
\end{figure*}
\begin{figure}[h]
	\centering
	\includegraphics[width=1.0\linewidth]{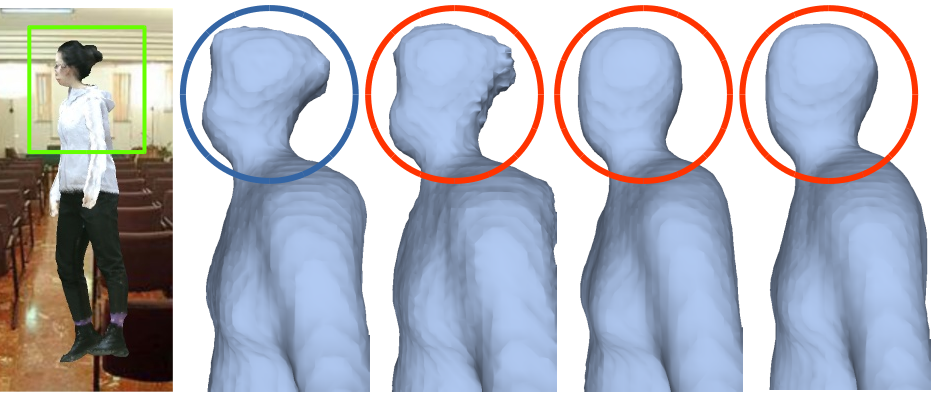}
	\caption{Visual evaluation of multi-scale volumetric feature transformation (VFT). From left to right: input image, head reconstruction result by our method, baseline (A), baseline (B) and baseline (C). }
	\label{fig:vft_eval_eval}
\end{figure}

\noindent\textbf{Baseline. } To evaluate our multi-scale VFT component, we implement 3 baseline networks: Baseline (A) only performs volumetric feature transformation at the finest scale, while Baseline (B) performs transformation at the coarest scale; different from the original network and Baseline (A)(B), Baseline (C) first encodes input images/volumes into latent codes, concatenates the latent code of the image with that of the volume and then feeds the concatenation into the volume decoder.  

\noindent\textbf{Results. } 
Fig.\ref{fig:vft_eval_loss} shows the reconstruction loss for different fusing methods. Here, we found by using multi-scale VFT, the network outperforms the baseline method in terms of the reconstruction of the model boundaries (see the second plot in Fig.\ref{fig:vft_eval_loss}). The same conclusion can be drawn from the visual comparison shown in Fig.\ref{fig:vft_eval_eval}. Using coarsest VFT (Baseline (B)) or latent code concantenation (Baseline (C)) results into over-smooth reconstruction of the girl's head due to the lack of higher-scale information (see the last two results in Fig.\ref{fig:vft_eval_loss}). The result generated by Baseline (A) is much more accurate but contain noises. With the proposed multi-scale VFT component, our network is able to reconstruct the hair bun of the girl (the blue circle in Fig.\ref{fig:vft_eval_loss}).

\subsubsection{Normal Refinement}
\begin{table}[h]
	\caption{Numerical normal errors with/without normal refinement. }
	\centering
	\begin{tabular}{l|c|c}
		\hline
		Error Metric & Cosine Distance & $\ell2$-norm \\
		\hline
		Without Refinement & 0.0941 & 0.336 \\
		With Refinement  & 0.0583 & 0.262 \\
		\hline
	\end{tabular}
	\label{tab:nml_eval_num}
\end{table}
\begin{figure}[h]
	\centering
	\subfigure[]{
		\centering
		\includegraphics[width=0.30\linewidth]{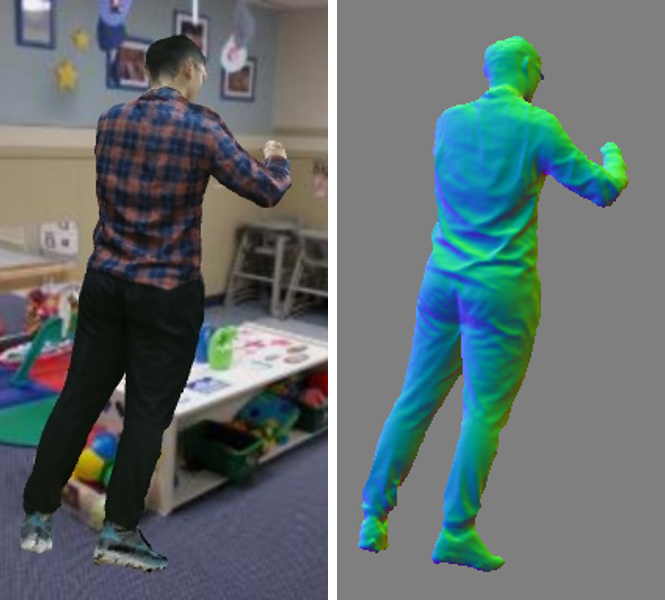}}
	\subfigure[]{
		\centering
		\includegraphics[width=0.30\linewidth]{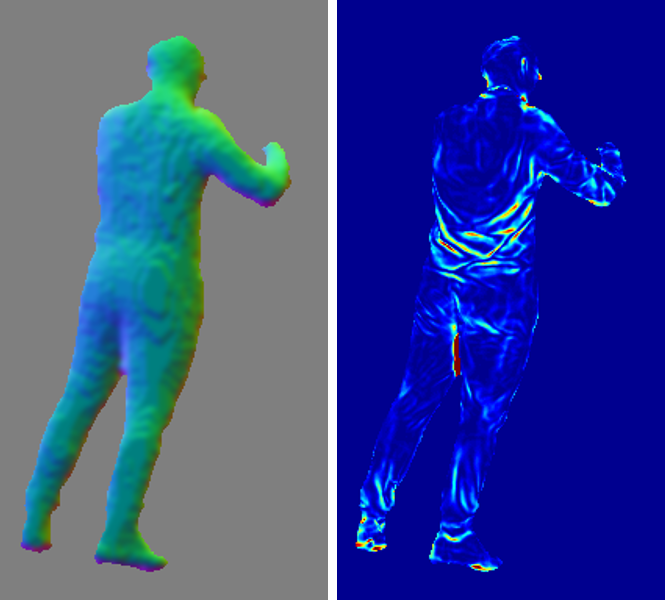}}
	\subfigure[]{
		\centering
		\includegraphics[width=0.30\linewidth]{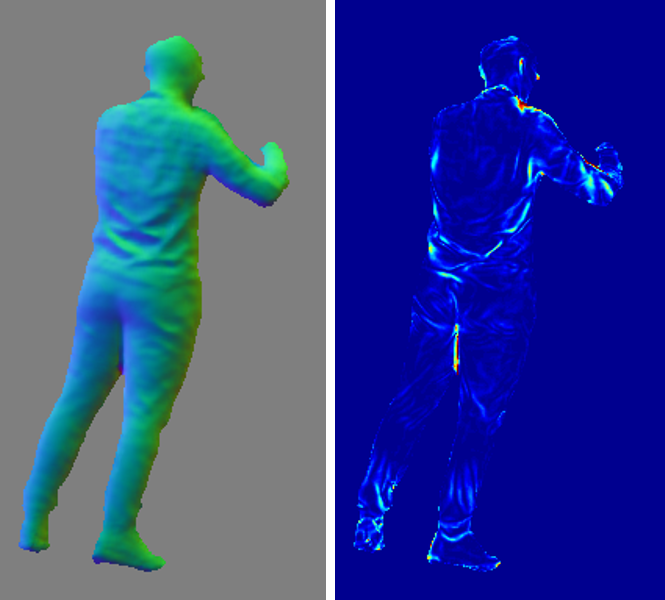}}
	\caption{Qualitative evaluation of normal refinement. (a) Reference image and the ground-truth normal. (b) Surface normal and error map without normal refinement. (c) Refined normal and the corresponding error map. }
	\label{fig:nml_eval_vis}
\end{figure}

\noindent\textbf{Baseline. } To evaluate our normal refinement module, we implement a baseline network by removing the volume-to-normal projection layer and the normal refinement U-Net as well from the original network.  

\noindent\textbf{Results. } The evaluation experiment is conducted using our synthetic dataset and the results are shown in Tab.\ref{tab:nml_eval_num} and Fig.\ref{fig:nml_eval_vis}. In Tab.\ref{tab:nml_eval_num} we present the prediction error of surface normal with and without normal refinement. This numeric comparison shows that the normal refinement network properly refines the surface normal based on the input image. We can also observe that surface details are enhanced and enriched after normal refinement in Fig.\ref{fig:nml_eval_vis}.  

\section{Discussion}
\begin{figure}
	\centering
	\includegraphics[width=0.8\linewidth]{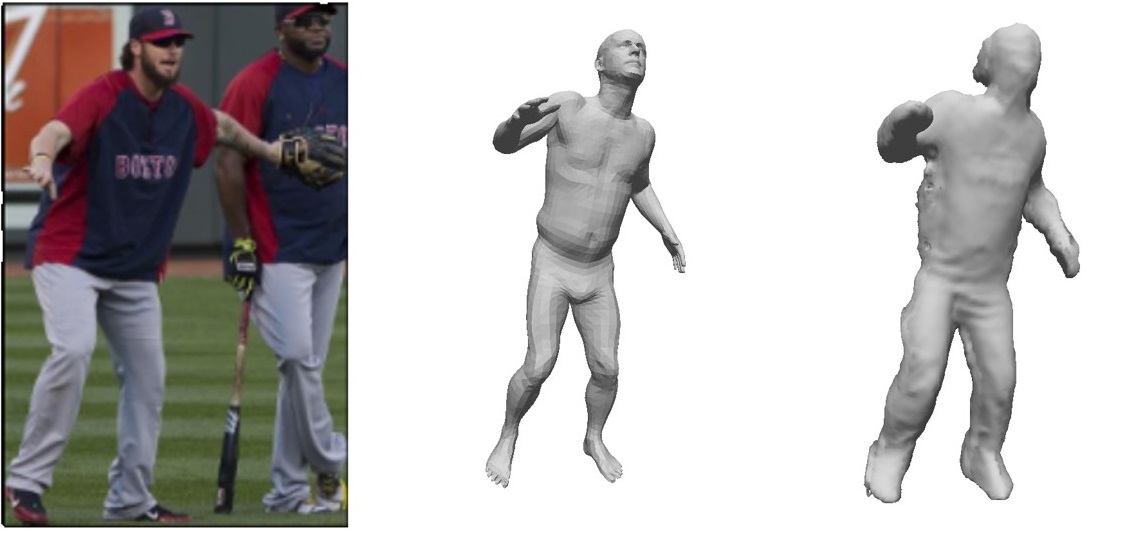}
	\caption{A failure case. HMR gives wrong prediction of the subject's upper body pose (middle), which results into wrong reconstruction by our network (right). }
	\label{fig:failure}
\end{figure}

\noindent\textbf{Limitations.} Our method relies on HMR to generate a dense semantic representation from SMPL model. Consequently, we cannot give an accurate reconstruction if the estimation of SMPL model is erroneous; see Fig.\ref{fig:failure} for an example. Additionally, the reconstruction of invisible areas is over-smoothed; using a generative adversarial network may force the network to learn to add realistic details to these areas. Our method also fails to recover fine-scale details such as facial expression and hands' motion. This issue can be addressed using methods that focus on face/hand reconstruction. 

\noindent\textbf{Conclusion.} In this paper, we have presented a deep-learning based framework to reconstruct a 3D human model from a single image. Based on the three-stage task decomposition, the dense semantic representation, the proposed network design and the 3D real-world human dataset, our method is able to estimate a plausible geometry of the target in the input image. We believe both our dataset and network will enable convenient VR/AR content creation and inspire further research on 3D vision for humans.

{\small
\bibliographystyle{ieee_fullname}
\bibliography{egpaper_final}
}

\end{document}